\documentclass[a4paper,conference]{IEEEtran}
\ifCLASSINFOpdf
\else
\fi

\usepackage[svgnames]{xcolor} 
\xdefinecolor{tucol1}{rgb}{1.0 0.5 0.05}
\xdefinecolor{tucol2}{rgb}{0.52,0.72,0.10}
\xdefinecolor{tucol3}{rgb}{0.0 0.0 0.8}
\xdefinecolor{tucol4}{rgb}{0.8 0.8 0}
\xdefinecolor{tucolv}{rgb}{0.1 0.35 0.17}
\xdefinecolor{tucol5}{rgb}{0.8 0 0.8}
\xdefinecolor{tucol6}{rgb}{0 0.8 0.8}
\xdefinecolor{tucol7}{rgb}{0.7 0.8 0.1}
\xdefinecolor{tucol8}{rgb}{0.2 0.7 0.34}
\xdefinecolor{tucol9}{rgb}{0.7 0.8 0.3}
\xdefinecolor{softgreen}{rgb}{0.49,0.50,0.47}

\usepackage{amsmath}
\usepackage{amsfonts}
\usepackage{hyperref}
\usepackage[pdftex]{graphicx} 
\usepackage[font={footnotesize}]{caption}
\usepackage{subcaption} 
\usepackage{tikz}
\usepackage{pgfplots}
\usepackage{algorithm}
\usepackage[noend]{algpseudocode}
\usepackage{multirow}
\usepackage{multicol}

\begin{document}
%
\title{Learning Attribute Representation for Human Activity Recognition}


\author{\IEEEauthorblockN{Fernando Moya Rueda, Gernot A. Fink}
\IEEEauthorblockA{TU Dortmund University\\Department of Computer Science\\Dortmund, Germany\\ \{fernando.moya, gernot.fink\}@tu-dortmund.de}}


%


\maketitle

\begin{abstract}
    Attribute representations became relevant in image recognition and word spotting, providing support under the presence of unbalance and disjoint datasets. However, for human activity recognition using sequential data from on-body sensors, human-labeled attributes are lacking. This paper introduces a search for attributes that represent favorably signal segments for recognizing human activities. It presents three deep architectures, including temporal-convolutions and an IMU centered design, for predicting attributes. An empiric evaluation of random and learned attribute representations, and as well as the networks is carried out on two datasets, outperforming the state-of-the art.
\end{abstract}


%
\IEEEpeerreviewmaketitle


\section{Introduction}

    Human activity recognition (HAR) is a classification task for recognizing human movements. It is the core of smart assistive technologies, e.g., in smart-homes, in rehabilitation and health support, and in the industry \cite{grzeszick2017_DNNBHAROPP}. HAR uses as inputs signals from videos or a set of on-body sensors. This paper covers HAR tasks using multichannel time series signals acquired from a set of on-body sensors. The recognition of human activities is a complicated task due to the large intra- and inter-class variability of human actions. Humans carry out the same tasks in different ways; even, a single person realizes a task differently. Furthermore, HAR is difficult due to the unbalance problem, where there are more samples of certain actions than others. Based on the assumption that body movements present certain patterns, HAR's idea is to classify them with different techniques.
    
    Multichannel time-series based HAR uses a combination of signals recorded from different types of sensors, e.g., accelerometers, gyroscopes, magnetometers, and heart rate monitors \cite{ronao2015_DCNNHARSS,ordonez2016_DCLRNNMWAR, hammerla2016_DCRMHARUW}. Usually, segmentation by means of a sliding window, extraction of engineered-features followed by a classification constitute the standard pipeline for recognizing human actions \cite{feldhorst2016_MCAOPPMS}. Nevertheless, relevant features are hard to compute, not accurate, and not scalable. Convolutional neural networks (CNNs) have been successfully applied to HAR tasks, unifying conveniently the feature extraction and classification \cite{yang2015_DCNNMTSHAR}. These architectures extract hierarchically the basic and complex features of the human body movements and learn their temporal dependencies.
    
    Motivated by the success of attribute representations for image classification \cite{lampert2009_LDUOCBCAT}, human actions can be likewise represented by a collection of attributes. These attributes describe semantically and coarsely human actions, i.e., attributes like moving left foot and right foot, forward, and sequential can be taken as the "walking" action \cite{zheng2017_SASVR}. Common attributes represent a set of similar human actions. For example, "walking" and "running" could have the movement of the feet as common attributes. The usage of this representation is suitable for recognition tasks where the data is unbalanced or training and testing sets are disjoint, e.g., zero shot learning. As such collection of attributes in the context of multichannel time series-based HAR is not available, we propose learning an attribute representation for HAR using deep architectures.
    
    The paper is structured as follows: \autoref{sec:related} will discuss the related work in the field of multichannel time series based HAR. In \autoref{sec:attr}, attribute representations in images and sequences will be introduced. In \autoref{sec:network} and \autoref{sec:random}, deep architectures and learning of attribute representations are described. Experiments on two datasets, the Opportunity- gestures and locomotion datasets, and the Pamap2 datasets, will be presented in \autoref{sec:datasets} and \autoref{sec:experiments}. In the last section, conclusions will be drawn

\section{Related Work} \label{sec:related}

    Traditionally, HAR using multichannel time series from on-body sensors has been solved by using engineered features obtained by statistical processes in a sliding window manner. The mean, median, min, max and magnitude area are examples of statistical features. Classifiers, e.g., SVM, Random Forest, Dynamic Time Warping or Hidden Markov Models use these features for predicting action classes of windowed sequences \cite{feldhorst2016_MCAOPPMS}. These approaches work relatively well, when data is scarce and highly unbalanced. However, identifying relevant features is time consuming, leading to difficulty in scaling up activity recognition of complex high level behaviours \cite{ordonez2016_DCLRNNMWAR}.
    
    As deep neural networks have become the state-of-the-art in different tasks, e.g., image classification \cite{Krizhevsky2012-ICD}, image segmentation \cite{long2015_FCNSS,badrinarayanan2015_SEGNET}, speech recognition and word spotting \cite{almazan2014_WSREA,Sudholt2016-PAD}, they haven been, recently, deployed in HAR in a sliding window framework \cite{grzeszick2017_DNNBHAROPP,ronao2015_DCNNHARSS,ordonez2016_DCLRNNMWAR,hammerla2016_DCRMHARUW,yang2015_DCNNMTSHAR}. Deep neural networks allow to learn the features and the classifier in an end-to-end manner directly from the raw data of multichannel time series. CNNs exploit the hierarchical composition of human movements, where complex activities are a combination of basic movements. The authors in \cite{ronao2015_DCNNHARSS} proposed an architecture with temporal-convolutions applied over all sensors simultaneously. Their architecture is composed of two or three temporal-convolution layers with ReLU activation functions followed by a downsampling (max-pooling), and a softmax classifier. The authors in \cite{yang2015_DCNNMTSHAR} introduced a deeper network with four temporal-convolution layers over individual sensors followed by a fully-connected layer and a softmax classifier. The fully-connected layer finds correlations between different sensors. An architecture that combines temporal-convolutions and Long Short-Term Mermory (LSTM) units is presented by the authors in \cite{ordonez2016_DCLRNNMWAR}. LSTMs are recurrent units with memory cells and a gating system that find long temporal dependencies in time-series problems without the limitations of exploiting or vanishing gradients during learning \cite{ordonez2016_DCLRNNMWAR,hochreiter1997_LSTM}. Their architecture is composed of four temporal-convolution layers with ReLU activation functions followed by two LSTM layers and a softmax classifier. While temporal-convolutions are applied to individual sensors de-noising and capturing local dependencies, the LSTM layers find temporal dependencies over all the pre-convolved sensor sequences.
    
    The authors in \cite{grzeszick2017_DNNBHAROPP} introduced a CNN based on IMUs for HAR in the context of the Order Picking process. The network has parallel convolutional blocks, one per IMUs' data. These blocks find intermediate representations of the IMUs' sequences, which are then concatenated by means of fully-connected layers. This network becomes robust against IMU's faults and asynchronous data. The authors in \cite{hammerla2016_DCRMHARUW} compared different architectures including CNNs, similar to \cite{yang2015_DCNNMTSHAR}, and LSTMs. They used a three-layered LSTM and one-layered bi-directional LSTM (B-LSTM) configurations directly on the raw data from the time-series. The B-LSTM processes the input sequences following two directions (forward, and backward). This network shows the state-of-the-art performance.
    

\section{Attribute representation} \label{sec:attr}

    Attributes provide high level semantic descriptions of objects, categories, and scenes in images \cite{lampert2009_LDUOCBCAT,almazan2014_WSREA,zheng2017_SASVR}. For example, in object classification, attributes can be colour, shape, texture, or size of objects. In HAR, collection of verbs and objects have been used for representing human actions in images and videos \cite{zheng2017_SASVR,yao2011_HARLBAAP}. Attribute representations have been used for zero-shot and transfer learning on different tasks like object recognition \cite{lampert2009_LDUOCBCAT} and word spotting \cite{almazan2014_WSREA,Sudholt2016-PAD}. The advantage of using an attribute representation is that recognition tasks can be carried out without much annotations, or in cases where data are highly unbalanced, the quantity of samples is large, and the testing set contains unseen object classes at training \cite{lampert2009_LDUOCBCAT}.
    
\begin{figure}[!h]
        \centering

        \begin{tikzpicture}[scale=0.8]
            \tikzstyle{circle1}=[circle, draw=black, minimum size=0.5cm, line width=0.2mm, inner sep=0pt]
            \tikzstyle{circle2}=[circle,draw=black, minimum size=0.09cm, line width=0.1mm, inner sep=0pt, fill=black]
            \tikzstyle{arrow1} = [->, line width=0.02cm, draw=black]
            
            \node [] (c01) at (-2.2,-1.0) {$\omega_{y_1}$};
            \node [] (c02) at (-2.8,-2.0) {$\omega_{y_2}<\omega_{y_1}$};
            
            \node [circle1] (c01) at (-1.5,-1.0) {$x_1$};
            \node [circle1] (c02) at (-1.5,-2.0) {$x_2$};
            
            \node [circle1] (c1) at (0.0,0.0) {$a_1$};
            \node [circle1] (c2) at (0.0,-1.0) {$a_2$};
            \node [circle2] (c3) at (0.0,-1.7) {};
            \node [circle2] (c4) at (0.0,-2.0) {};
            \node [circle2] (c5) at (0.0,-2.3) {};
            \node [circle1] (c6) at (0.0,-3.0) {$a_3$};
            
            \draw  [arrow1](c01) edge (c1);
            \draw  [arrow1](c01) edge (c2);
            \draw  [arrow1](c01) edge (c6);
            
            \draw  [arrow1](c02) edge (c1);
            \draw  [arrow1](c02) edge (c2);
            \draw  [arrow1](c02) edge (c6);
            
            \node [circle1] (c7) at (1.5,-0.5) {$y_1$};
            \node [circle1] (c8) at (1.5,-1.2) {$y_2$};
            \node [circle2] (c9) at (1.5,-1.8) {};
            \node [circle2] (c10) at (1.5,-2.0) {};
            \node [circle2] (c11) at (1.5,-2.2) {};
            \node [circle1] (c12) at (1.5,-2.7) {$y_3$};
            
            \draw  [arrow1](c1) edge (c7);
            \draw  [arrow1](c1) edge (c8);
            \draw  [arrow1](c1) edge (c12);

            \draw  [arrow1](c2) edge (c7);
            \draw  [arrow1](c2) edge (c8);
            \draw  [arrow1](c2) edge (c12);
            
            \draw  [arrow1](c6) edge (c7);
            \draw  [arrow1](c6) edge (c8);
            \draw  [arrow1](c6) edge (c12);

        \end{tikzpicture}

        \caption{Graphical representation of the HAR using attribute representations as an intermediate layer. HAR using attribute representation could be enhanced as classes with more samples, e.g., $y_{1}$ with $\omega_{y_1}$ could help learning lesser frequent classes, e.g., $y_{2}$.}
        \label{fig:mapping}
    \end{figure}
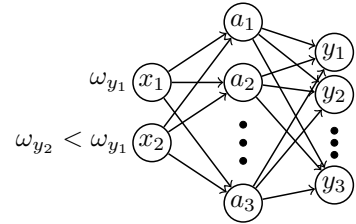

    Consider for HAR a dataset with tuples $(x_1, y_1),...,(x_n, y_n)$ with $X = x_1,x_2,...,x_n$ being $N$ sample sequences from an arbitrary space and $Y=y_1,y_2,..,y_n$ their respective classes from a set of $K$ number of classes, the idea is to learn the function $f: X \rightarrow Y$.  Using an attribute representation $A$, an additional mapping is introduced $f: X \rightarrow A \rightarrow Y$, see \autoref{fig:mapping}. This additional mapping serves as an intermediate layer that allows sharing high-level concepts among classes, making full usage of the whole data. Classes with higher amount of samples can borrow attributes to lesser frequent classes. 
    
    In the context of multichannel time-series based HAR, datasets are highly unbalanced, specially towards the "NULL" class. This class covers all the human actions that are not relevant for the task, sometimes involving more than $75\%$ of the recorded data \cite{ordonez2016_DCLRNNMWAR}. An attribute representation is beneficial for solving this case, where sequences of the "NULL" class provide good material for learning shared attributes that are contained in less frequent classes, see \autoref{fig:mapping}.

\section{Attribute-based CNN architecture} \label{sec:network}

    \begin{figure*}[!ht]
        \centering

        \begin{tikzpicture} [x=1.0cm,y=0.9cm]
    
            \tikzstyle{node1}=[text=black, font=\normalsize \bfseries];
            \tikzstyle{node2}=[text=black, font=\small \bfseries];
            \tikzstyle{node3}=[text=black, font=\small];
            \tikzstyle{node4}=[text=black, font=\footnotesize];
            \tikzstyle{arrow1} = [line width=0.05]
            \tikzstyle{circle1}=[circle,draw=black, minimum size=0.1cm, line width=0.2mm, inner sep=0pt]
            \tikzstyle{circle2}=[circle,draw=black, minimum size=0.05cm, line width=0.1mm, inner sep=0pt, fill=black]

            \node [node1] at (0.0-2.5,0.9){IMU 1};
            
            \node (label) at (0.0 - 1.1, 0.0 + 0.9)[draw=white, line width=0.0]{
                    \includegraphics[width= 0.075\textwidth]{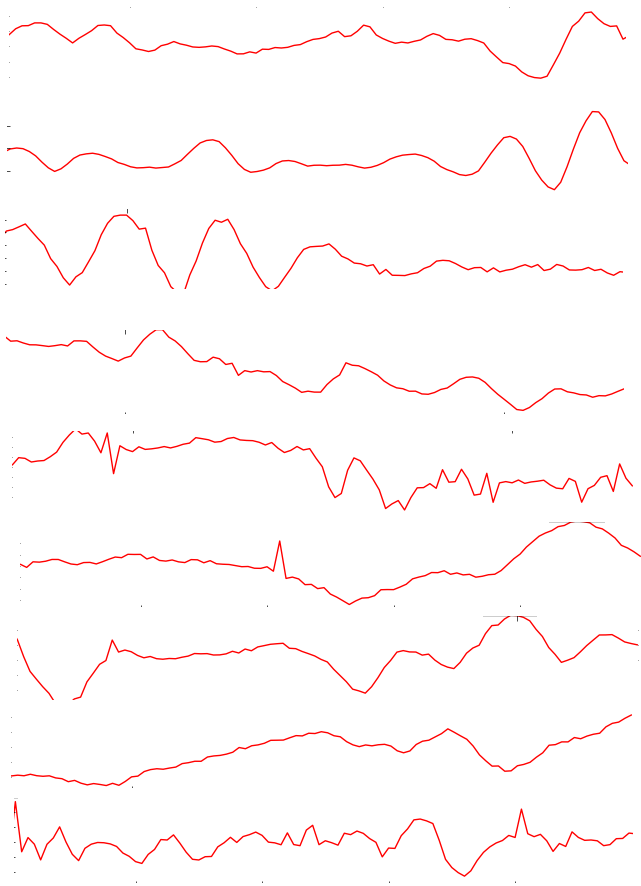}
                  }; 
                  
            \draw [line width=0.05mm, opacity=0.4] (0.0-1.8,0.0 - 0.1) rectangle +(1.4,2.0);
            
            \node [node3] at (0.0+0.5,0.0-0.3){$C=64$};
            
            \draw [line width=0.05mm,fill=tucol4, opacity=0.3](1.4,1.8)--(0.4,1.8)--(0,1.5)--(1,1.5)--(1.4,1.8)--(1.4,0.3)--(1,0)--(0,0)--(0,1.5)--(1,1.5)--cycle;
            \draw  [arrow1](1,0.0) edge (1,1.5);

            \node [node3] at (0.0+0.5+1.5,0.0-0.3){$C=64$};
            
            \draw [line width=0.05mm,fill=tucol4, opacity=0.3](1.4 + 1.5,1.8)--(0.4 + 1.5,1.8)--(0 + 1.5,1.5)--(1 + 1.5,1.5)--(1.4 + 1.5,1.8)--(1.4 + 1.5,0.3)--(1 + 1.5,0)--(0 + 1.5,0)--(0 + 1.5,1.5)--(1 + 1.5,1.5)--cycle;
            \draw  [arrow1](1 + 1.5,0.0) edge (1 + 1.5,1.5);

            \draw [line width=0.05mm,fill=tucol1, opacity=0.3](1.2 + 3,1.8)--(0.4 + 3,1.8)--(0 + 3,1.5)--(0.8 + 3,1.5)--(1.2 + 3,1.8)--(1.2 + 3,0.3)--(0.8 + 3,0)--(0 + 3,0)--(0 + 3,1.5)--(0.8 + 3,1.5)--cycle;
            \draw  [arrow1](0.8 + 3,0.0) edge (0.8 + 3,1.5);

            \node [node3] at (0.0+0.5+4.3,0.0-0.3){$C=64$};
            
            \draw [line width=0.05mm,fill=tucol4, opacity=0.3](1.2 + 4.3,1.8)--(0.4 + 4.3,1.8)--(0 + 4.3,1.5)--(0.8 + 4.3,1.5)--(1.2 + 4.3,1.8)--(1.2 + 4.3,0.3)--(0.8 + 4.3,0)--(0 + 4.3,0)--(0 + 4.3,1.5)--(0.8 + 4.3,1.5)--cycle;
            \draw  [arrow1](0.8 + 4.3,0.0) edge (0.8 + 4.3,1.5);
            
            \node [node3] at (0.0+0.5+5.6,0.0-0.3){$C=64$};
            
            \draw [line width=0.05mm,fill=tucol4, opacity=0.3](1.2 + 5.6,1.8)--(0.4 + 5.6,1.8)--(0 + 5.6,1.5)--(0.8 + 5.6,1.5)--(1.2 + 5.6,1.8)--(1.2 + 5.6,0.3)--(0.8 + 5.6,0)--(0 + 5.6,0)--(0 + 5.6,1.5)--(0.8 + 5.6,1.5)--cycle;
            \draw  [arrow1](0.8 + 5.6,0.0) edge (0.8 + 5.6,1.5);

            \draw [line width=0.05mm,fill=tucol1, opacity=0.3](1.0 + 6.9,1.8)--(0.4 + 6.9,1.8)--(0 + 6.9,1.5)--(0.6 + 6.9,1.5)--(1.0 + 6.9,1.8)--(1.0 + 6.9,0.3)--(0.6 + 6.9,0)--(0 + 6.9,0)--(0 + 6.9,1.5)--(0.6 + 6.9,1.5)--cycle;
            \draw  [arrow1](0.6 + 6.9,0.0) edge (0.6 + 6.9,1.5);
            
            \node [node4] at (0.0+0.5+8.8,0.0+1.4){$C=512$};
            
            \draw [fill=tucol2, line width=0.05mm, opacity=0.4] (8.5,0.0) rectangle +(0.2,1.8);

            \node [circle2] (c5) at (0.0-1.0,1.0-1.5) {};
            \node [circle2] (c6) at (0.0-1.0,0.8-1.5) {};
            \node [circle2] (c7) at (0.0-1.0,0.6-1.5) {};


            \node [node1] at (0.0-2.5,0.9-3.0){IMU m};
            
            \node (label) at (0.0 - 1.1, 0.0 + 0.9 - 3.0)[draw=white, line width=0.0]{
                    \includegraphics[width= 0.075\textwidth]{images/seq_conv1.png}
                  }; 
                  
            \draw [line width=0.05mm, opacity=0.4] (0.0-1.8,0.0 - 3.1) rectangle +(1.4,2.0);

            \draw [line width=0.05mm,fill=tucol4, opacity=0.3](1.4,1.8- 3.0)--(0.4,1.8 - 3.0)--(0,1.5 - 3.0)--(1,1.5 - 3.0)--(1.4,1.8 - 3.0)--(1.4,0.3- 3.0)--(1,0- 3.0)--(0,0- 3.0)--(0,1.5- 3.0)--(1,1.5- 3.0)--cycle;
            \draw  [arrow1](1,0.0- 3.0) edge (1,1.5- 3.0);

            \draw [line width=0.05mm,fill=tucol4, opacity=0.3](1.4 + 1.5,1.8- 3.0)--(0.4 + 1.5,1.8- 3.0)--(0 + 1.5,1.5- 3.0)--(1 + 1.5,1.5- 3.0)--(1.4 + 1.5,1.8- 3.0)--(1.4 + 1.5,0.3- 3.0)--(1 + 1.5,0- 3.0)--(0 + 1.5,0- 3.0)--(0 + 1.5,1.5- 3.0)--(1 + 1.5,1.5- 3.0)--cycle;
            \draw  [arrow1](1 + 1.5,0.0- 3.0) edge (1 + 1.5,1.5- 3.0);

            \draw [line width=0.05mm,fill=tucol1, opacity=0.3](1.2 + 3,1.8- 3.0)--(0.4 + 3,1.8- 3.0)--(0 + 3,1.5- 3.0)--(0.8 + 3,1.5- 3.0)--(1.2 + 3,1.8- 3.0)--(1.2 + 3,0.3- 3.0)--(0.8 + 3,0- 3.0)--(0 + 3,0- 3.0)--(0 + 3,1.5- 3.0)--(0.8 + 3,1.5- 3.0)--cycle;
            \draw  [arrow1](0.8 + 3,0.0- 3.0) edge (0.8 + 3,1.5- 3.0);
            
            \draw [line width=0.05mm,fill=tucol4, opacity=0.3](1.2 + 4.3,1.8- 3.0)--(0.4 + 4.3,1.8- 3.0)--(0 + 4.3,1.5- 3.0)--(0.8 + 4.3,1.5- 3.0)--(1.2 + 4.3,1.8- 3.0)--(1.2 + 4.3,0.3- 3.0)--(0.8 + 4.3,0- 3.0)--(0 + 4.3,0- 3.0)--(0 + 4.3,1.5- 3.0)--(0.8 + 4.3,1.5- 3.0)--cycle;
            \draw  [arrow1](0.8 + 4.3,0.0- 3.0) edge (0.8 + 4.3,1.5- 3.0);

            \draw [line width=0.05mm,fill=tucol4, opacity=0.3](1.2 + 5.6,1.8- 3.0)--(0.4 + 5.6,1.8- 3.0)--(0 + 5.6,1.5- 3.0)--(0.8 + 5.6,1.5- 3.0)--(1.2 + 5.6,1.8- 3.0)--(1.2 + 5.6,0.3- 3.0)--(0.8 + 5.6,0- 3.0)--(0 + 5.6,0- 3.0)--(0 + 5.6,1.5- 3.0)--(0.8 + 5.6,1.5- 3.0)--cycle;
            \draw  [arrow1](0.8 + 5.6,0.0- 3.0) edge (0.8 + 5.6,1.5- 3.0);
            
            \draw [line width=0.05mm,fill=tucol1, opacity=0.3](1.0 + 6.9,1.8- 3.0)--(0.4 + 6.9,1.8- 3.0)--(0 + 6.9,1.5- 3.0)--(0.6 + 6.9,1.5- 3.0)--(1.0 + 6.9,1.8- 3.0)--(1.0 + 6.9,0.3- 3.0)--(0.6 + 6.9,0- 3.0)--(0 + 6.9,0- 3.0)--(0 + 6.9,1.5- 3.0)--(0.6 + 6.9,1.5- 3.0)--cycle;
            \draw  [arrow1](0.6 + 6.9,0.0- 3.0) edge (0.6 + 6.9,1.5- 3.0);

            \draw [fill=tucol2, line width=0.05mm, opacity=0.4] (8.5,0.0- 3.0) rectangle +(0.2,1.8);

            
            \draw [line width=0.05mm, fill=tucol8, opacity=0.1](6.7+2.0,1.8)--(6.7+2.0,-0.3-2.7)--(6.7+2.5,-0.2-1.3)--(6.7+2.5,1.6-1.3)--cycle;
            
            \node [node4] at (0.0+0.5+8.0,0.0-0.5){concat.};
            
            \draw [fill=tucol8, line width=0.05mm, opacity=0.4] (8.7 + 0.5,0.0- 1.5) rectangle +(0.2,1.8);
            
            \node [node4] at (0.0+10.0,0.0+0.3){$C=512$};
            
            \draw [fill=tucol2, line width=0.05mm, opacity=0.4] (8.7 + 1.0,0.2- 1.5) rectangle +(0.2,1.4);
            
            \draw [fill=tucol5, line width=0.05mm, opacity=0.4] (8.7 + 1.7,0.2- 1.5) rectangle +(0.2,1.4);

            \draw [fill=tucol4, line width=0.05mm, opacity=0.4] (8.7 + 0.5,0.2- 2.8) rectangle +(0.2,0.2);
            
            \node [node2] at (8.7 + 2.3,0.2- 2.7){[$5\times1$] temporal-conv};
            
            \draw [fill=tucol1, line width=0.05mm, opacity=0.4] (8.7 + 0.5,0.2- 3.2) rectangle +(0.2,0.2);
            
            \node [node2] at (8.7 + 2.2,0.2- 3.1){[$2\times1$] max-pooling};
            
            \draw [fill=tucol5, line width=0.05mm, opacity=0.4] (8.7 + 0.5,0.2- 3.6) rectangle +(0.2,0.2);
            
            \node [node2] at (8.7 + 2.2,0.2- 3.5){Sigmoid act-function};
                
        \end{tikzpicture}

        \caption{The attrCNN-IMU architecture contains $m$ parallel temporal convolutional blocks, one per IMU. The ouputs of the blocks are concatenated and forwarded to a fully connected layer. The output layer is the sigmoid function. The attrCNN and the attrCNNLSTM have only one block for all of the IMUs, and they have fully connected and LSTM layers respectivelly.}
        \label{fig:networks}
    \end{figure*}

    In this paper, we proposed three attribute-based architectures based on the CNN, deepConvLSTM presented in \cite{ordonez2016_DCLRNNMWAR, yang2015_DCNNMTSHAR}, and the CNN-IMU introduced in \cite{grzeszick2017_DNNBHAROPP}. These architectures have in common temporal-convolution layers and fully-connected layers. The convolutional layers extract temporal local features providing and creating an abstract representation of the input sequence. The fully-connected units connect all the local features giving a global view of the input data. Specifically, the CNN architecture contains four temporal-convolution and two fully-connected layers with ReLU activation functions. The deepConvLSTM has also four temporal-convolution layers with ReLU activation functions, but it uses two LSTM layers instead of the fully-connected ones. The LSTMs capture the global temporal dynamics of the input data. The CNN-IMU network contains parallel temporal-convolution blocks, each with an additional fully-connected layer, see \autoref{fig:networks}. These parallel branches process and merge input sequences from IMUs individually. They create an intermediate representation per IMU. The network combines these intermediate representations into a global one, forwarding them to a softmax layer. Depending on the dataset, the architectures contain also max-pooling layers. For all of the three architectures, each of the convolutional layers has $C=64$ filters of size $[5,1]$ performing convolutions only in the time-axis. Both the fully-connected layer and the LSTMs layer contain $128$ units. Max-pooling of $[2,1]$ with stride of $1$ is used.
    
    The network's input consists of sequence segments of length $T$ measurements composed of multiple sensor channels $D$ extracted from a sliding window approach with step size $s$. So, the input's size is $[T,D]$. Having a feature map $x^{i}$ of size $[T,D,C]$ in layer $i$, and a set of $C_{i}$ filters $w^{i,c_{i}}$ of size $[F,1]$ and biases $b^{i,c_{i}}$ connecting layers $i$ and $j$, a temporal-convolution for each $d$ sensor is:
    
    \begin{equation}
        \begin{matrix}
            x_{t,d}^{j,c_{j}}=\sigma \left (  \sum \limits_{c=0}^{C_{i}}\sum\limits_{f=0}^{F-1}w_{f}^{i,c} \cdot  x_{t+f,d}^{i,c} +b^{i,c} \right ) &
             & \forall d = 1,...,D
        \end{matrix}
    \end{equation}
    
    where $\sigma$ is the activation function. The filters $w^{i}$ are shared among the sensors $D$.
    
    A max-pooling operation between layer $i$ and $j$ for a single channel $c$ reduces the size of a feature map finding the maximum among a set of $P$ values, and is given by:
    
    \begin{equation}
        \begin{matrix}
            x_{t,d}^{(j),c_{j}} = \max\limits_{0< p\leq P} \left (   x_{t+p,d}^{i,c_{i}}  \right )   & \forall d=1,...,D
        \end{matrix}
    \end{equation}

    A standard CNN for classification uses a softmax activation, see \autoref{softmax}, which produces pseudo-probabilities per class $k_i \in K$. For training, the cross-entropy loss between the estimated probabilities $x^{j}_{k}$, and the target label $y_k \in Y$ is used.
    
    \begin{multicols}{2}
        \begin{equation} \label{softmax}
            x^{j}_{k}=\frac{e^{x^{i} } }{\sum_{k=1}^{K} e^{x^{i}_{k}} }
        \end{equation}\break
        \begin{equation} \label{sigmoid}
            sigmoid(x) = \frac{1}{1+e^{-x}}
        \end{equation}
    \end{multicols}
    
    However, the proposed network will compute an attribute representation of an input sequence, rather than classify it using a softmax function, being different from \cite{ordonez2016_DCLRNNMWAR} and \cite{grzeszick2017_DNNBHAROPP}. As the representation will contain multiple $0s$ and $1s$ for an attribute being present or not, the softmax function is replaced by a sigmoid activation function, see \autoref{sigmoid}, following the architecture described in \cite{Sudholt2016_PAD} in the context of word spotting. The sigmoid activation function is applied to each element of the output layer. Its output corresponds to pseudo-probabilities for each attribute $a_{i}$ being present in the representation.
    
    Such a network must be trained using the binary-cross entropy loss given by.
    
    \begin{equation} \label{eq:binary-cross}
        E(a,\tilde{a}) = -\frac{1}{n}\sum _{i=1}^{n} \left ( a_ilog\; \tilde{a_{i}} + (1-a_i)log\; (1-\tilde{a_{i}}) \right )
    \end{equation}
    
    where $a$ is the target attribute representation and $\tilde{a}$ is the CNN's output.
    
    \autoref{fig:networks} shows the CNN-IMU architecture for attribute representation, called attrIMU-CNN. The other two networks (attrCNN and attrCNNLSTM) are simplifications of the first one. Both are a branch of the CNN-IMU, where their inputs are sequences including all the sensors from the IMUs. The attrdeepCNNLSTM contains LSTMs layers instead of fully-connected ones.


\section{Learning Attribute Representations} \label{sec:random}
    
    In the context of HAR using multichannel time-series signals and having introduced the attribute representation and the attribute-based deep networks, datasets do not possess annotated attribute representations. Their annotations are related with specific coarse actions, e.g., walk, jump, run. However, there is no annotation of small and granulate actions that could describe a coarse one. For example, for walking, a person should move first one foot and then the other in a coordinate way and speed, and for running these movements are faster. For standing, the person does not move any of its foot. One might think of discriminating walking, running and standing by the attributes moving or not moving, the left or right foot, and the speed. Nevertheless, these annotations do not exist. Besides, humans can not provide easily annotations by just observing the data, as for example they can on images or videos.
    
    Considering the literature in random sub-space projection and random indexing for representing words, phrases and documents  \cite{sahlgren2005_AIRI,kanerva1994_SCECML,paradis2013_FSETURIV}, a set of different sequence segments could be projected to, or described by a random representation. This random representation should be adequate such that one can differentiate the actions. As this random space is spanned by unknown attributes, it might not be suitable. One needs to search for the best possible representation that will describe properly a set of sequence segments. An evolutionary algorithm can be used for performing this search, where attribute representations are seen as genotypes of the classes. By evaluating, selecting and mutating those genotypes, a proper representation of the classes can be learned.
    
    Algorithm \ref{alg:evol} shows the evolutionary algorithm (EA) used for finding the best attribute representation of a certain dataset. The algorithm evaluates the fitness of an attribute representation by training and validating a deep network. The validation weighted $F1$ serves as the performance metric, in which case, first, the cosine distance between the computed $\tilde{a}$ and $a\in A$ target attributes is estimated, finding class predictions. Second, the precision and recall of the predictions are used, see \autoref{eq:f1}. The algorithm mutates, evaluates and selects the best attribute representation for a certain number of iterations starting from a random representation. A global mutation in $\mathbb{B}^n$ is selected to change the attribute representation; that is, the attribute $a_i$ for $i=1,...,n$ of a single target attribute representation $a$ flips with probability $p_i \in (0,1)$.

    \begin{algorithm}
        \scriptsize
        \caption{Evolutionary algorithm}\label{alg:evol}
        \begin{algorithmic}[1]
            \Procedure{Evolution}{$niter,K, tr, vl$}\Comment{Input niter:number of iterations, K: number of classes, tr: training set, vl: validation set}
            \State $A_{gen} \gets random(K)$
            \State $F_1^{best} \gets 0.0$
            \For {$i=1 : niter$}\Comment{Evaluate each generation}
                \State $network \gets trainCNN(tr, A_{gen})$
                \State $F_{1}^{i} \gets testCNN(network, vl, A_{gen})$
                \If{$F_{1}^{i}>F_{1}^{best}$}
                    \State $A_{best} \gets A_{gen}$
                    \State $F_{1}^{best} \gets F_{1}^{i}$
                \EndIf
                \State $A_{gen} \gets mutate(A_{gen})$
            \EndFor\label{evolutiondFor}
            \State \textbf{return} $A_{best}$\Comment{Best attribute generation}
            \EndProcedure
        \end{algorithmic}
    \end{algorithm}
    
    As shown in \autoref{sec:datasets}, the datasets are highly unbalanced and the classification accuracy is not a convenient measurement for evaluating the performance of the networks. The weighted $F1$ considers the correct classification of each class equally, using the precision, recall and the proportion of a class in the dataset \cite{ordonez2016_DCLRNNMWAR}. 
    
    \begin{equation}\label{eq:f1}
        F_1 = \sum_i 2 \times \frac{n_i}{N} \times \frac{precision_i \times recall_i}{precision_i + recall_i}
    \end{equation}
    
    where,  $n_i$ is the number of samples for each class $k\in K$, and $N$ is the total number of samples in the dataset.
    
    The evolution is performed for a fix number of iterations $niter$ in which a single attribute generation is used for training a network. The network is trained from scratch within each iteration, i.e., weights from previous generations are not used. The performance on the validation set is taken as the generation's fitness. The best generation or learned attribute set is used for a final training and deploying on a testing set.

\section{Datasets} \label{sec:datasets}

    An evaluation of the proposed approach is carried out for HAR on the datasets: Opportunity \cite{chavarriaga2013_TOC}, and Pamap2 \cite{reiss2012_IANBDAM}. In general, the performance of three CNNs using a learned attribute representations is evaluated. First, A proper attribute representation for each dataset and the networks is found by means of the evolutionary algorithm, see \autoref{sec:random}. Second, Using this learned representation, a final training of the networks is done.
    
    \subsection{Opportunity}
        The Opportunity dataset \cite{roggen2010_CCADHRNSE, chavarriaga2013_TOC} contains recordings of fours participants carrying out activities of daily living (ADL). Seven IMUs, placed on both wrists, upper arms, shoes, and back, provide recordings of $113$ sensors. Each participant executed five sessions of ADLs. There were no restrictions on performing the activities. In addition, participants carried out a drill session with a predefined sorted set of $17$ activities. The dataset contains two different label annotations: \textbf{Gestures} with $K=18$ classes and \textbf{Locomotion} with $K=5$ classes. Therefore, one obtains two different classification tasks. The gestures task has, for example, activities like open and closing doors or drawers, and drinking coffee, and the locomotion has activities related with motion like walking, standing, and sitting. Following \cite{ordonez2016_DCLRNNMWAR, chavarriaga2013_TOC}, sessions ADL$3$ of participants $2$ and $3$ are used as validation set, and sessions ADL$4$ and ADL$5$ of participants $2$ and $3$ as testing set. The recording's sample rate is $30Hz$. A sliding window of $720$ms or $T=24$ and a step size of $360$ms or $s=12$ ($50\%$ overlapping as \cite{ordonez2016_DCLRNNMWAR}) are used for segmenting the sequences.

    \subsection{Pamap2}
        The Pamap2 dataset \cite{reiss2012_IANBDAM} contains recordings from $9$ participants carrying out $12$ activities, e.g., walking, standing, cycling. Three IMUs (on the hand, chest and ankle) and a HR-monitor provide recordings of $40$ sensors, e.g., accelerometers, gyroscope, magnetometer, temperature and heart rate. Recordings from participants $5$ and $6$ are used as validation and testing sets respectively. Following \cite{hammerla2016_DCRMHARUW},  IMUs recordings are downsampled to $30Hz$. A sliding window of $3$s or $T=100$ and step size of $660ms$ or $s=22$ ($78\%$ overlapping as in \cite{hammerla2016_DCRMHARUW, reiss2012_IANBDAM}) are used for segmenting the sequences. The window size is smaller than the one in \cite{hammerla2016_DCRMHARUW} for generating a larger number of segments.
    
    \subsection{Preprocessing}
    
        As suggested in \cite{ordonez2016_DCLRNNMWAR,grzeszick2017_DNNBHAROPP}, input sequences were normalized per sensor to the range $[0,1]$. Moreover, a Gaussian noise of $\mu=0$ and $\sigma=0.01$ is added, simulating sensors' inaccuracies.


\section{Experiments} \label{sec:experiments}

    \begin{figure*}[!ht]
        \centering
        \begin{subfigure}[b]{0.33\textwidth}
            \begin{tikzpicture}[xscale=0.7,yscale=0.7]
                \begin{axis}[ 
                    xmode=log,
                    title={\normalsize{Pamap2 Dataset}},
                    xlabel={Evolution iteration [log]},
                    ylabel={Weighted $F_1[\%]$},
                    xmin=1, xmax=500,
                    ymin=75, ymax=91,
                    xtick={1,5, 10,50,100,200, 500},
                    ytick={75,80,85,90,91},
                    extra y ticks={50},
                    legend style={at={(0.6,0.2)},anchor=west},
                    ymajorgrids=true,
                    grid style=dashed,
                    ]
                    
                    \tikzstyle{node1}=[text=blue, font=\scriptsize];
                 
                    \addplot[color=blue, mark=*, nodes near coords, every node near coord/.append style={font=\tiny}]
                        coordinates {
                        (1,16.6)(2,82.7)(4,83.6)(6,83.7)(7,84.7)(15,85.5)(16,87.2)(143,87.3)(216,87.8)(341,88.3)(426,89.2)
                        };
                    \addplot[color=tucol2, mark=*, nodes near coords, every node near coord/.append style={font=\tiny}]
                        coordinates {
                        (1,86.5)(3,87.78)(75,88.5)(325,88.63)
                        };
                    \addplot[color=red, mark=*, nodes near coords, every node near coord/.append style={font=\tiny}]
                        coordinates {
                        (0,77.56)(1,78.71)(2,84.00)(3,88.50)(249,88.98)
                        };
                      
                    \node[node1] at (axis cs: 3,76) {(1,16.6)};    
                 
                \end{axis}
    
            \end{tikzpicture}
        \end{subfigure}
        \hspace{-0.0em}
        \begin{subfigure}[b]{0.33\textwidth}

            \begin{tikzpicture}[xscale=0.7,yscale=0.7]
                \begin{axis}[
                    xmode=log,
                    title={\normalsize{Gestures Dataset}},
                    xlabel={Evolution iteration [log]},
                    xmin=1, xmax=2500,
                    ymin=86, ymax=91,
                    xtick={1,10,100,300,800,2200},
                    ytick={86, 87, 88, 89, 90, 91},
                    legend style={at={(0.6,0.2)},anchor=west},
                    ymajorgrids=true,
                    grid style=dashed,
                    ]
                 
                    \addplot[color=blue, mark=*, nodes near coords, every node near coord/.append style={font=\tiny}]
                        coordinates {
                        (1,89.5)(8,89.6)(56,89.7)(135,89.8)(683,89.82)(1989,90.0)
                        };
                    \addplot[color=tucol2, mark=*, nodes near coords, every node near coord/.append style={font=\tiny}]
                        coordinates {
                        (1,88.23)(5,88.24)(7,88.34)(17,88.52)(25,88.66)(33,88.78)(66,88.8)(119,89.1)(545,89.3)
                        };
                    \addplot[color=red, mark=*, nodes near coords, every node near coord/.append style={font=\tiny}]
                        coordinates {
                        (1,87.57)(3,87.72)(4,89.33)(79,90.01)
                        };
                 
                \end{axis}
            \end{tikzpicture}
            
        \end{subfigure}
        \hspace{-2em}
        \begin{subfigure}[b]{0.33\textwidth}

            \begin{tikzpicture}[xscale=0.7,yscale=0.7]
                \begin{axis}[
                    xmode=log,
                    title={\normalsize{Locomotion Dataset}},
                    xlabel={Evolution iteration [log]},
                    xmin=1, xmax=2000,
                    ymin=83, ymax=89,
                    xtick={1,10,100,300,800,2000},
                    ytick={83, 84, 85, 86, 87, 88, 89},
                    legend style={at={(0.6,0.2)},anchor=west, font = \footnotesize},
                    ymajorgrids=true,
                    grid style=dashed,
                    ]
                 
                    \addplot[color=blue, mark=*, nodes near coords, every node near coord/.append style={font=\tiny}]
                        coordinates {
                        (1,86.38)(260,86.42)(280,86.59)(333,86.78)(464,87.31)(1140,87.45)
                        };
                    \addplot[color=tucol2, mark=*, nodes near coords, every node near coord/.append style={font=\tiny}]
                        coordinates {
                        (1,84.57)(5,85.56)(7,85.65)(14,85.81)(38,85.86)(50,86.55)(199,86.59)(202,86.70)(237,86.93)(313, 86.95)
                        };
                    \addplot[color=red, mark=*, nodes near coords, every node near coord/.append style={font=\tiny}]
                        coordinates {
                        (1,86.34)(2,86.47)(61,86.86)(132,87.0)(166,87.44)(168,87.78)(528,88.14)
                        };
                    \legend{attrCNN, attrCNN-IMU, attrLSTM}
                 
                \end{axis}
            \end{tikzpicture}
        
        \end{subfigure}

        \caption{Weighted $F1[\%]$ of the attributes evolution for the three attrCNNs on the three datasets.}
        \label{fig:evol_validation}
    \end{figure*}
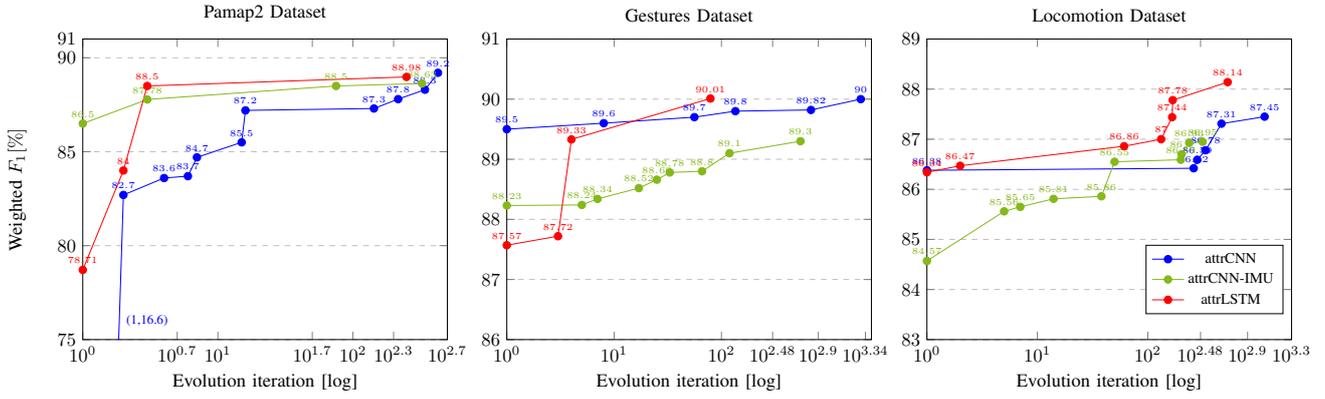

    


    The algorithm \ref{alg:evol} finds an attribute representation $A_{best}$ suitable for each dataset, by iteratively training and validating the three architectures using attribute configurations, which mutate starting from random. Different numbers $n$ of attributes $\in \mathbb{B}^n$ for representing a class $k$ are proposed depending on the number of total classes $K$ per dataset. Based on training and validating the architectures on random attributes for different $n$, we set $n=[10,32,24]$ number of attributes for the Opportunity-Locomotion, -Gestures datasets, and Pamap2 dataset. Depending on the dataset and network, the following configurations have been employed. The parameters are updated by minimizing the binary-cross entropy, see \autoref{eq:binary-cross}, using batch gradient descent with RMSProp update rule \cite{ordonez2016_DCLRNNMWAR}. We used a learning rate of $0.0001$, RMS decay of $0.9$, and batch size of $100$ for Gestures and Locomotion and of $50$ for Pamap2. In addition, we used Dropout on the feature-maps inputs of the first, and second fully-connected layers. Networks are trained from scratch for a fixed number of epochs, see \autoref{tab:epochs}, in each EA-iteration. A nearest neighbour approach was used for predicting the class $k$ by measuring the cosine distance from the attribute vectors $\tilde{a}$ to the set $a\in A_{best}$. We used the validation weighted $F1$ as the fitness metric.
    
    \begin{table}[!h]
        \scriptsize
        \caption{Number of epochs for training the architectures on the evolutionary algorithm.}
        \label{tab:epochs}
        \centering
        \begin{tabular}{c|c|c|c}
            \multirow{2}{*}{\textbf{Dataset}} & \multicolumn{3}{c}{\textbf{Network}} \\
            \cline{2-4}
             &  attrCNN & attrCNN-IMU & attrdeepConvLSTM \\
            \hline
            \hline
            Gestures    & 12 & 5 & 10\\
            Locomotion  & 10 & 5 & 10\\
            Pamap2      & 25 & 5 & 25\\
        \end{tabular}
    \end{table}
    
    \autoref{fig:evol_validation} shows the weighted F$1$ evolution for the three networks on the three validation datasets. The weighted $F1$ on the Pamap2 presents the biggest improvements of $72.65\%$, $2.51\%$, $11.42\%$ using the attrCNN, attrCNN-IMU, and attrdeepCNNLSTM between the first and the last attribute generation in $niter=[426, 75, 249]$ iterations respectively. The weighted $F1$ on the Gestures and Locomotion shows average improvements of $1.39\%$ and $2.39\%$ respectively. It is noticeable that mostly the performance of the networks present a relative good performance despite the randomness of the initial attribute generations. This could be explained as the Gestures and Locomotion datasets are strongly unbalanced towards the NULL class, and they predict correctly sequence segments with that label. However, as the attributes evolve, the correct predictions increase, showing the potential of the attribute search. Moreover, common attributes among classes are found, which helps sharing strength between the more and the less frequent classes. In addition, there is a substantial dimensionality reduction.

    \autoref{tab:attributes} shows the learned attribute representation for attrCNN on the Opportunity dataset with Locomotion labels after $1140$ iterations. Analyzing the attribute sharing among action classes, there is a sort of semantic relation. For example, the "Stand" class shares $4$ and $5$ attributes with classes "Sit" and "Lie". These three action classes have lesser sharing with "Walk" class with $3$ and $2$ shared attributes. One advantage of this attribute sharing is that the "Null" class attributes become a source for learning attributes shared with infrequent classes. There is not a specific definition of each attribute, but the sort of basic movements or states in each action class keeps certain relation. One uses these relations for learning a better representation of the actions. Observing the attributes found using the attrCNN-IMU on the same dataset, the classes "Stand" and "Walk" shared more attributes than with the class "Sit". These examples show that the learned attribute representation depicts a certain logic, but it changes with respect to the networks.

    \begin{table}[!h]
        \scriptsize
        \caption{Attribute representation $A \in \mathbb{B}^{10}$ of the Locomotion dataset found using the evolutionary algorithm and the attrCNN}
        \label{tab:attributes}
        \centering
        \begin{tabular}{c|c|c|c|c|c|c|c|c|c|c}
            \multirow{2}{*}{\textbf{Classes}} & \multicolumn{10}{c}{\textbf{Attributes}} \\
            \cline{2-11}
             &  1 & 2 & 3 & 4 & 5 & 6 & 7 & 8 & 9 & 10 \\
            \hline
            \hline
            Null    & 1 & 0 & 0 & 0 & 0 & 1 & 0 & 0 & 1 & 1\\
            Stand   & 0 & 1 & 1 & 0 & 1 & 1 & 1 & 1 & 1 & 1\\
            Walk    & 0 & 1 & 0 & 0 & 0 & 0 & 0 & 0 & 1 & 1\\
            Sit     & 0 & 0 & 0 & 0	& 1 & 1 & 0 & 0 & 1 & 1\\
            Lie     & 1 & 1 & 1 & 0 & 1 & 0 & 0 & 1 & 1 & 0
        \end{tabular}
    \end{table}

    The learned attribute representation for Gesture dataset using the three networks show multiple relations. As the dataset contains mostly opening and closing of doors, fridges, dishwashers, and drawers, the attributes are mixed. These labels also explains in part the good performance of the initial random representation utilizing the three networks on this dataset. However, after the attribute evolution, classes involving only closing or only opening movements display a strong attribute sharing with $20-22$ common attributes.
    
    For the Pamap2 dataset, the learned attribute representation for the networks presents different relations. This dataset has different activities, including moving and static ones, but without the "Null" class. The classes are more diverse, having more marked action classes. So, the shared attribute are also more diverse, e.g., it keeps relations among classes "rope jumping", "lying", "cycling", "ironing".
    
    \subsection{Final training}
    
        Finally, we evaluate the performance of the networks using the learned attribute representations on the testing sets. We have trained the three attrCNNs on the training and validation sets using the learned attribute representations. Similar settings on learning rates, momentum, and updating rule from the evolution are used. However, the number of epochs is increased. In addition, random attributes were used for training and testing the attrCNNs for $niter=100$. In that case, we report the average of their weighted $F1[\%]$ performances. \autoref{tab:finalExp} presents the testing weighted $F1$.

        \begin{table}[!h]
            \scriptsize
            \caption{Weighted $F1[\%]$ of the three attrCNNs on the testing sets using the evolved and random attribute representations. In addition, it shows a comparison with the state-of-the-art networks. Results marked with '*' were obtained following the networks and configurations of the original publications.}
            \label{tab:finalExp}
            \centering
            \begin{tabular}{l|c|c|c}
                \textbf{Architecture} &  \textbf{Pamap2} & \textbf{Locomotion} & \textbf{Gestures}\\
                \hline
                \hline
                CNN \cite{ordonez2016_DCLRNNMWAR} & 87.37* & 87.8 & 85.1\\
                deepCNNLSTM \cite{ordonez2016_DCLRNNMWAR} & 87.63* & 89.5 & 91.5\\
                CNN \cite{hammerla2016_DCRMHARUW} & 87.2* & - & 90.8\\
                CNN-IMU \cite{grzeszick2017_DNNBHAROPP} & 89.01* & 88.23* & 92.15* \\
                \hline
                attrCNN random & 84.72 & 85.9 & 88.96  \\
                attrdeepCNNLSTM random & 85.15 & 86.64 & 88.58 \\
                attrCNN-IMU random & 86.26 & 86.85 & 89.92 \\
                \hline
                attrCNN evol & 90.55 & \textbf{90.0} & 91.94\\
                attrdeepCNNLSTM evol & 88.0 & 89.01 & 90.97 \\
                attrCNN-IMU evol & \textbf{90.88} &  89.75 & \textbf{92.9}
            \end{tabular}
        \end{table}

        It is noticeable the comparable performance of these representations. Deep networks are capable of learning features and classify input segments when the targets are represented by a set of attributes either random or learned. The attrCNN's and the attrCNN-IMU's performances using the learned attributes are higher in comparison with the random ones. These results demonstrate that the attributes found by means of the evolutionary algorithm on the validation set keep a relation with action classes, and represent better the input sequences. Moreover, these performances also show that the networks benefit from the shared attributes, using information from the most frequent actions reducing the effects of the unbalanced problem. The attrCNN and the attrCNN-IMU using the learned attributes show better performance with respect to similar architectures with action labels as targets from \cite{ordonez2016_DCLRNNMWAR, grzeszick2017_DNNBHAROPP}. The attrdeepCNNLSTMs do not benefit greatly in comparison with the deepCNNLSTM; however, their performances are comparable.


\section{Conclusion} \label{sec:conclusion}

    We have presented a new application of attribute representations for solving human activity recognition from multichannel time-series signals acquired from on-body sensors. Considering that human-labeled attributes annotations for HAR are not existing, we proposed a method for learning attributes that better represent the time-series signals, starting from a random representation. Three deep architectures for mapping input sequences into a set of attributes for recognition are introduced. We show empirically that these networks using either random or learned attributes, in general, present a comparable or even better performance contrasting with similar networks that predict classes directly.







%

\bibliographystyle{IEEEtran}
\bibliography{literatur}

\begin{thebibliography}{10}
\providecommand{\url}[1]{#1}
\csname url@samestyle\endcsname
\providecommand{\newblock}{\relax}
\providecommand{\bibinfo}[2]{#2}
\providecommand{\BIBentrySTDinterwordspacing}{\spaceskip=0pt\relax}
\providecommand{\BIBentryALTinterwordstretchfactor}{4}
\providecommand{\BIBentryALTinterwordspacing}{\spaceskip=\fontdimen2\font plus
\BIBentryALTinterwordstretchfactor\fontdimen3\font minus
  \fontdimen4\font\relax}
\providecommand{\BIBforeignlanguage}[2]{{%
\expandafter\ifx\csname l@#1\endcsname\relax
\typeout{** WARNING: IEEEtran.bst: No hyphenation pattern has been}%
\typeout{** loaded for the language `#1'. Using the pattern for}%
\typeout{** the default language instead.}%
\else
\language=\csname l@#1\endcsname
\fi
#2}}
\providecommand{\BIBdecl}{\relax}
\BIBdecl

\bibitem{grzeszick2017_DNNBHAROPP}
R.~Grzeszick, J.~M. Lenk, F.~M. Rueda, G.~A. Fink, S.~Feldhorst, and M.~ten
  Hompel, ``Deep neural network based human activity recognition for the order
  picking process,'' in \emph{In Proc. of the 4th Int. Workshop on Sensor-based
  Activity Recognition and Interaction}.\hskip 1em plus 0.5em minus 0.4em\relax
  ACM, 2017.

\bibitem{ronao2015_DCNNHARSS}
C.~A. Ronao and S.-B. Cho, ``Deep convolutional neural networks for human
  activity recognition with smartphone sensors,'' in \emph{Int. Conf. on Neural
  Information Processing}.\hskip 1em plus 0.5em minus 0.4em\relax Springer,
  2015, pp. 46--53.

\bibitem{ordonez2016_DCLRNNMWAR}
F.~J. Ord{\'o}{\~n}ez and D.~Roggen, ``Deep convolutional and lstm recurrent
  neural networks for multimodal wearable activity recognition,''
  \emph{Sensors}, vol.~16, no.~1, p. 115, 2016.

\bibitem{hammerla2016_DCRMHARUW}
N.~Y. Hammerla, S.~Halloran, and T.~Ploetz, ``Deep, convolutional, and
  recurrent models for human activity recognition using wearables,''
  \emph{arXiv preprint arXiv:1604.08880}, 2016.

\bibitem{feldhorst2016_MCAOPPMS}
S.~Feldhorst, M.~Masoudenijad, M.~ten Hompel, and G.~A. Fink, ``Motion
  classification for analyzing the order picking process using mobile
  sensors,'' in \emph{In Proc. of the 5th Int. Conf. on Pattern Recognition
  Applications and Methods}.\hskip 1em plus 0.5em minus 0.4em\relax
  SCITEPRESS-Science and Technology Publications, Lda, 2016, pp. 706--713.

\bibitem{yang2015_DCNNMTSHAR}
J.~Yang, M.~N. Nguyen, P.~P. San, X.~Li, and S.~Krishnaswamy, ``Deep
  convolutional neural networks on multichannel time series for human activity
  recognition.'' in \emph{IJCAI}, 2015, pp. 3995--4001.

\bibitem{lampert2009_LDUOCBCAT}
C.~H. Lampert, H.~Nickisch, and S.~Harmeling, ``Learning to detect unseen
  object classes by between-class attribute transfer,'' in \emph{Computer
  Vision and Pattern Recognition, 2009. CVPR 2009. IEEE Conference on}.\hskip
  1em plus 0.5em minus 0.4em\relax IEEE, 2009, pp. 951--958.

\bibitem{zheng2017_SASVR}
J.~Zheng, Z.~Jiang, and R.~Chellappa, ``Submodular attribute selection for
  visual recognition,'' \emph{IEEE transactions on pattern analysis and machine
  intelligence}, vol.~39, no.~11, pp. 2242--2255, 2017.

\bibitem{Krizhevsky2012-ICD}
A.~Krizhevsky, I.~Sutskever, and G.~E. Hinton, ``{Imagenet classification with
  deep convolutional neural networks},'' in \emph{Advances in neural
  information processing systems}, 2012, pp. 1097--1105.

\bibitem{long2015_FCNSS}
J.~Long, E.~Shelhamer, and T.~Darrell, ``Fully convolutional networks for
  semantic segmentation,'' in \emph{In Proc. of the IEEE Conference on Computer
  Vision and Pattern Recognition}, 2015, pp. 3431--3440.

\bibitem{badrinarayanan2015_SEGNET}
V.~Badrinarayanan, A.~Handa, and R.~Cipolla, ``Segnet: A deep convolutional
  encoder-decoder architecture for robust semantic pixel-wise labelling,''
  \emph{arXiv preprint arXiv:1505.07293}, 2015.

\bibitem{almazan2014_WSREA}
J.~Almaz{\'a}n, A.~Gordo, A.~Forn{\'e}s, and E.~Valveny, ``Word spotting and
  recognition with embedded attributes,'' \emph{IEEE transactions on pattern
  analysis and machine intelligence}, vol.~36, no.~12, pp. 2552--2566, 2014.

\bibitem{Sudholt2016-PAD}
S.~Sudholt and G.~A. Fink, ``{PHOCNet}: {A} deep convolutional neural network
  for word spotting in handwritten documents,'' in \emph{Proc. Int. Conf. on
  Frontiers in Handwriting Recognition}, 2016.

\bibitem{hochreiter1997_LSTM}
S.~Hochreiter and J.~Schmidhuber, ``Long short-term memory,'' \emph{Neural
  computation}, vol.~9, no.~8, pp. 1735--1780, 1997.

\bibitem{yao2011_HARLBAAP}
B.~Yao, X.~Jiang, A.~Khosla, A.~L. Lin, L.~Guibas, and L.~Fei-Fei, ``Human
  action recognition by learning bases of action attributes and parts,'' in
  \emph{Computer Vision (ICCV), IEEE Int. Conf}.\hskip 1em plus 0.5em minus
  0.4em\relax IEEE, 2011, pp. 1331--1338.

\bibitem{Sudholt2016_PAD}
S.~Sudholt and G.~A. Fink, ``{PHOCNet: A Deep Convolutional Neural Network for
  Word Spotting in Handwritten Documents},'' in \emph{Proc. Int. Conf. on
  Frontiers in Handwriting Recognition}, Shenzhen, China, 2016.

\bibitem{sahlgren2005_AIRI}
M.~Sahlgren, ``An introduction to random indexing,'' in \emph{Methods and
  Applications of Semantic Indexing Workshop at the 7th Int. Conf. on
  Terminology and Knowledge Engineering, TKE 2005}, 2005.

\bibitem{kanerva1994_SCECML}
P.~Kanerva, ``The spatter code for encoding concepts at many levels,'' in
  \emph{ICANN’94}.\hskip 1em plus 0.5em minus 0.4em\relax Springer, 1994, pp.
  226--229.

\bibitem{paradis2013_FSETURIV}
R.~D. Paradis, J.~K. Guo, J.~Moulton, D.~Cameron, and P.~Kanerva, ``Finding
  semantic equivalence of text using random index vectors,'' \emph{Procedia
  Computer Science}, vol.~20, pp. 454--459, 2013.

\bibitem{chavarriaga2013_TOC}
R.~Chavarriaga, H.~Sagha, A.~Calatroni, S.~T. Digumarti, G.~Tr{\"o}ster,
  J.~d.~R. Mill{\'a}n, and D.~Roggen, ``The opportunity challenge: A benchmark
  database for on-body sensor-based activity recognition,'' \emph{Pattern
  Recognition Letters}, vol.~34, no.~15, pp. 2033--2042, 2013.

\bibitem{reiss2012_IANBDAM}
A.~Reiss and D.~Stricker, ``Introducing a new benchmarked dataset for activity
  monitoring,'' in \emph{Wearable Computers (ISWC), 2012 16th Int. Symposium
  on}.\hskip 1em plus 0.5em minus 0.4em\relax IEEE, 2012, pp. 108--109.

\bibitem{roggen2010_CCADHRNSE}
D.~Roggen, A.~Calatroni, M.~Rossi, T.~Holleczek, K.~F{\"o}rster,
  G.~Tr{\"o}ster, P.~Lukowicz, D.~Bannach, G.~Pirkl, A.~Ferscha \emph{et~al.},
  ``Collecting complex activity datasets in highly rich networked sensor
  environments,'' in \emph{Networked Sensing Systems (INSS), 2010 Seventh Int.
  Conf. on}.\hskip 1em plus 0.5em minus 0.4em\relax IEEE, 2010, pp. 233--240.

\end{thebibliography}


%

\end{document}